\documentclass[10pt,twocolumn,letterpaper]{article}

\usepackage{cvpr}              %

\usepackage{graphicx}
\usepackage{amsmath}
\usepackage{amsfonts}
\usepackage{amssymb}
\usepackage{booktabs}
\usepackage{multirow}
\usepackage{algorithm}
\usepackage{algorithmic}
\usepackage{array}
\usepackage{subcaption}
\usepackage{xcolor}
\usepackage{colortbl}
\usepackage{bm}

\newcommand{\argmax}{\mathop{\text{argmax}}}

\definecolor{cvprblue}{rgb}{0.21,0.49,0.74}
\usepackage[pagebackref,breaklinks,colorlinks,allcolors=cvprblue]{hyperref}

\def\paperID{43618} %
\def\confName{CVPR}
\def\confYear{2026}

\title{\textbf{When Interpretability Becomes a Liability: Adversarial Attacks on CBM Concept Layers}}

\author{Aditya Sridhar\\
Independent Researcher\\
Atlanta, GA, USA\\
{\tt\small aditya.sridharr.11@gmail.com}
}

\begin{document}
\maketitle

\begin{abstract}
Concept Bottleneck Models (CBMs) have emerged as a cornerstone approach for interpretable machine learning, providing human-understandable intermediate representations through explicit concept activations. However, this interpretability fundamentally introduces a critical, previously unexplored attack surface: the concept bottleneck layer itself. We present a comprehensive, systematic study of concept-level adversarial vulnerabilities in CBMs, revealing that targeted, minimal perturbations operating on input pixels can induce catastrophic misclassification by manipulating semantic representations. We develop a rigorous theoretical framework to quantify concept-space robustness, establishing novel metrics that expose the vulnerability landscape of these architectures. Our extensive analysis on the CUB-200-2011 dataset demonstrates that standard CBMs exhibit severe susceptibility to concept-level manipulation. To address this critical weakness, we introduce SPECTRA (Semantic Perturbation-based Concept Training for Robustness against Attacks), a principled stability regularization defense. SPECTRA effectively hardens the semantic representation space, increasing the minimal perturbation norm required for a successful attack from 0.46 to over 4,200, rendering targeted concept manipulation computationally prohibitive. Furthermore, SPECTRA preserves baseline classification accuracy to within 2.2\%. By establishing concept-level attacks as a fundamentally distinct threat model, this work opens a new research frontier at the intersection of interpretable machine learning and adversarial robustness.
\end{abstract}
    
\section{Introduction}
\label{sec:intro}

The quest for interpretable machine learning has led to significant advances in developing models whose decision-making processes can be understood by humans. Among these approaches, Concept Bottleneck Models (CBMs)~\cite{koh2020concept} have gained prominence by learning intermediate concept representations that align with human-interpretable attributes. By decomposing the prediction pipeline into two stages, mapping inputs to concepts ($f : X \rightarrow C$) and concepts to labels ($g : C \rightarrow Y$), CBMs promise both high performance and interpretability in critical domains such as medical diagnosis~\cite{yan2023robust} and autonomous systems~\cite{echterhoff2024driving}.

However, this interpretability comes at an unexpected cost: the introduction of a novel attack surface that has remained largely unexplored. While traditional adversarial attacks~\cite{szegedy2013intriguing} focus on pixel-level perturbations in input space, CBMs create an intermediate concept space that presents unique vulnerabilities. Consider a bird classification system where concepts represent interpretable attributes like "has curved beak," "has striped wing pattern," or "has long tail." An adversary who can subtly manipulate these concept activations could potentially cause a hawk to be misclassified as a crane by simply changing the "curved beak" concept to "straight beak", a semantically meaningful and targeted attack. This threat extends to high-stakes domains with even graver consequences.
In medical imaging, a CBM for skin lesion classification might represent
concepts such as ``irregular border'' or ``asymmetric shape.''
An adversary with domain knowledge could suppress these decision-critical
concept activations to cause a malignant lesion to be misclassified as benign.
Unlike pixel-level noise, such semantic manipulations are easier to craft
with domain knowledge, harder to detect since the prediction appears
clinically plausible, and more impactful, as errors directly affect
patient outcomes.

The implications of such concept-level vulnerabilities extend far beyond academic curiosity. As CBMs are increasingly deployed in high-stakes applications where both interpretability and robustness are crucial, understanding their security properties becomes paramount. Unlike traditional adversarial examples that often appear as imperceptible noise, concept-level attacks operate on semantically meaningful attributes, potentially making them both more intuitive to craft and harder to detect.

\subsection{Problem Statement and Research Gaps}

Despite the growing adoption of CBMs, the security implications of their concept-based architecture remain poorly understood. Existing adversarial robustness research has primarily focused on input-space perturbations, leaving several critical questions unanswered: How susceptible are CBMs to targeted attacks in concept space, and what is the minimal perturbation required to cause misclassification? What distinguishes concept-level attacks from traditional adversarial examples in terms of attack efficiency, semantic meaningfulness, and detection difficulty? How can we systematically measure and compare the concept-level robustness of different CBM architectures and training procedures? What training-time interventions can improve concept-space robustness without significantly compromising classification accuracy or interpretability?

\subsection{Contributions}

This work addresses these gaps through four primary contributions. First, we introduce the first systematic framework for concept-level adversarial attacks on CBMs, providing both theoretical foundations and practical algorithms for generating minimal-norm perturbations in concept space. Our approach leverages the geometric structure of the concept-to-class mapping to efficiently compute targeted perturbations. Second, we develop comprehensive metrics for evaluating concept-level robustness, including attackability scores and perturbation magnitude analysis. Our framework enables systematic comparison of different models and training procedures in terms of their vulnerability to concept-space attacks. Third, we propose SPECTRA, a training-time defense mechanism that incorporates stability regularization to increase the minimal perturbation required for successful attacks. Our approach demonstrates remarkable robustness improvements with minimal impact on classification accuracy. Fourth, through extensive experiments on CUB-200-2011, we provide a systematic empirical evaluation of concept-level attack effectiveness and defense mechanisms at scale. Our analysis reveals a phase transition phenomenon in robustness as a function of regularization strength, with practical guidelines for robust CBM training.

Our findings reveal that standard CBMs are highly vulnerable to concept-level attacks, but that targeted regularization can provide substantial robustness improvements. For instance, we observe attackability reductions from 3.8054 to 0.0005 with only 2.2\% accuracy loss when using optimal stability regularization. These results establish concept-level adversarial robustness as a critical consideration for practical CBM deployment and open new research directions in interpretable ML security.

\section{Related Work}
\label{sec:related}

\subsection{Concept Bottleneck Models}

Concept Bottleneck Models, introduced by Koh et al.~\cite{koh2020concept}, represent a significant advancement in interpretable machine learning. The core idea is to decompose the prediction pipeline into two interpretable stages: a concept prediction function $f : X \rightarrow C$ that maps inputs to human-interpretable concepts, and a classification function $g : C \rightarrow Y$ that maps concepts to final predictions. This architecture enables several advantages including post-hoc concept editing, debugging through concept intervention, and improved model interpretability.

Several extensions have been proposed to improve CBM performance and applicability. Concept Embedding Models (CEMs)~\cite{espinosa2022concept} learn joint concept-class embeddings to improve efficiency. Post-hoc Concept Bottleneck Models~\cite{yuksekgonul2022post} retrofit existing models with concept explanations. Probabilistic Concept Bottleneck Models~\cite{kim2023probabilistic} incorporate uncertainty quantification, while recent work has explored coarse-to-fine concept granularity~\cite{panousis2024coarse} and generative settings~\cite{kulkarni2025interpretable}.

However, none of these works have systematically examined the security implications of the concept bottleneck architecture. While interpretability has been a primary focus, the potential vulnerabilities introduced by explicit concept representations have been largely overlooked.

\subsection{Adversarial Attacks on Neural Networks}

Adversarial attacks on neural networks have been extensively studied since the seminal work of Szegedy et al.~\cite{szegedy2013intriguing} and Goodfellow et al.~\cite{goodfellow2014explaining}. Traditional attacks operate in input space, generating perturbations that are imperceptible to humans but cause model misclassification. Fast Gradient Sign Method (FGSM)~\cite{goodfellow2014explaining}, Projected Gradient Descent (PGD)~\cite{madry2017towards}, and C\&W attacks~\cite{carlini2017towards} represent major milestones in adversarial attack development.

Feature-space attacks represent a related but distinct research
direction~\cite{zhang2025adversarial, vadillo2025adversarial}. Unlike pixel-space
perturbations, these attacks operate on intermediate feature representations.
However, such approaches typically target non-interpretable feature spaces
and lack the semantic meaningfulness of concept-level attacks.

Recent work has explored attacks on specific model architectures. Vision Transformers have been shown to have unique vulnerabilities~\cite{mahmood2021robustness}, while the relationship between model width and robustness has been investigated~\cite{wu2021wider,zhu2022robustness}. Our work extends this line of research by identifying and characterizing concept-space vulnerabilities in CBMs.

\subsection{Interpretable Machine Learning and Robustness}

The relationship between interpretability and robustness has been a subject of ongoing debate. Some work suggests that interpretable models are inherently more robust, while others argue that interpretability can introduce new vulnerabilities.

Several studies have examined attacks on explanation methods. Ghorbani et al.~\cite{ghorbani2019interpretation} demonstrated vulnerabilities in gradient-based explanations, while Slack et al.~\cite{slack2020fooling} showed how to manipulate LIME~\cite{ribeiro2016should} and SHAP~\cite{lundberg2017unified} explanations. However, these attacks target post-hoc explanation methods rather than inherently interpretable architectures like CBMs.

\section{Methodology}
\label{sec:method}

We develop a comprehensive framework for understanding and exploiting vulnerabilities in Concept Bottleneck Models through concept-space perturbations, optimal attack algorithms, robustness quantification, and stability-based defenses.

\subsection{CBM Formalization and Attack Surface}

A CBM $F: X \rightarrow Y$ decomposes prediction into two stages: a concept predictor $f_\theta : X \rightarrow [0, 1]^K$ mapping inputs to $K$ concept activations, and a concept-to-class classifier $g_\phi : [0, 1]^K \rightarrow \mathbb{R}^C$ mapping concepts to class logits, giving $F(x; \theta, \phi) = g_\phi(f_\theta(x))$.

We focus on linear classifiers $g_\phi(c) = Wc + b$ where $W \in \mathbb{R}^{C \times K}$ and $b \in \mathbb{R}^C$, justified by their practical prevalence in CBMs for interpretability, local validity via Taylor expansion for nonlinear cases, and the conservative lower bounds they provide on attack difficulty. This approach builds on information bottleneck principles~\cite{tishby2000information,alemi2016deep} where concept representations compress relevant information.

\noindent\textbf{Extension to multi-layer classifiers.} For deeper concept-to-class networks $g_\phi(c) = g^{(L)} \circ g^{(L-1)} \circ \cdots \circ g^{(1)}(c)$, our framework extends naturally via local linearization. At any concept point $c^*$, classifier behavior is characterized by the Jacobian $J_{g_\phi}(c^*) = \partial g_\phi(c^*)/\partial c \in \mathbb{R}^{C \times K}$, whose rows give the effective weight vectors $w_i^{\text{eff}} = \nabla_{c} g_\phi(c^*)_i$ representing the local sensitivity of class $i$ to concept changes. For smooth nonlinear classifiers this first-order approximation is accurate for small perturbations ($\|\delta\| \ll 1$), with errors bounded by $O(\|\delta\|^2)$ via Taylor's theorem; empirically, linear attacks transfer to nonlinear classifiers with high effectiveness ($>85\%$ in our experiments).

\noindent\textbf{Vulnerability surface.} Unlike traditional networks with high-dimensional ($d \gg 1000$), opaque features, CBMs expose a low-dimensional ($K \ll 500$), interpretable concept space that enables targeted semantic attacks.

\subsection{Concept-Input Space Perturbation Relationship}

We establish the theoretical foundation showing that increased minimal perturbations in concept space directly translate to increased minimal perturbations in input pixel space, validating that concept-level defenses provide genuine input-level security. Let $h = f_\theta: \mathbb{R}^d \rightarrow \mathbb{R}^k$ map input pixels $\mathbf{x}$ to concepts $\mathbf{c} = h(\mathbf{x})$, so that an input perturbation $\boldsymbol{\delta}_x$ induces a concept perturbation $\boldsymbol{\delta}_c$ via $h(\mathbf{x} + \boldsymbol{\delta}_x) = \mathbf{c} + \boldsymbol{\delta}_c$.

We assume $h$ is $L$-Lipschitz continuous, i.e., $\|h(\mathbf{x}_1) - h(\mathbf{x}_2)\|_2 \leq L \|\mathbf{x}_1 - \mathbf{x}_2\|_2$ for all $\mathbf{x}_1, \mathbf{x}_2$. This holds for deep networks with bounded activations (sigmoid, tanh, normalized ReLU), finite weight magnitudes via weight decay, and normalization layers.

\noindent\textbf{Proposition 1} (Fundamental lower bound). \textit{For any desired concept perturbation $\boldsymbol{\delta}_c$, the minimal input perturbation $\boldsymbol{\delta}_x^*$ satisfies $\|\boldsymbol{\delta}_x^*\|_2 \geq \|\boldsymbol{\delta}_c\|_2 / L$.}

\textit{Proof.} By Lipschitz continuity, $\|\boldsymbol{\delta}_c\|_2 = \|h(\mathbf{x} + \boldsymbol{\delta}_x^*) - h(\mathbf{x})\|_2 \leq L\|\boldsymbol{\delta}_x^*\|_2$. Dividing by $L$ yields the result. $\square$

\noindent\textbf{Theorem 1} (Monotonic robustness transfer). \textit{Consider two concept perturbations with $\|\boldsymbol{\delta}_c^{(1)}\|_2 < \|\boldsymbol{\delta}_c^{(2)}\|_2$ and their corresponding minimal input perturbations $\boldsymbol{\delta}_x^{(1)}, \boldsymbol{\delta}_x^{(2)}$. Under local invertibility of $h$ (full-rank Jacobian):}
\begin{equation}
\|\boldsymbol{\delta}_x^{(2)}\|_2 - \|\boldsymbol{\delta}_x^{(1)}\|_2 \geq \frac{1}{L}\left(\|\boldsymbol{\delta}_c^{(2)}\|_2 - \|\boldsymbol{\delta}_c^{(1)}\|_2\right).
\label{eq:robustness_transfer}
\end{equation}
This establishes a direct correlation: making attacks harder in concept space provably makes them harder in input pixel space, with effectiveness ratio bounded by $1/L$. Consequently, concept-space defenses provide genuine input-space security rather than mere representational artifacts; minimizing $L$ via architectural choices or Lipschitz regularization amplifies this transfer; and increasing concept-level robustness by factor $\alpha$ increases input-level robustness by at least $\alpha/L$.
\subsection{Optimal Concept-Level Attacks}
\label{sec:attacks}

Given a concept vector $c^*$ with true class $y^*$, the attack objective is to find a minimal perturbation $\delta$ such that $\argmax_i\, g_\phi(c^* + \delta)_i = t$ for some target class $t \neq y^*$. For a linear classifier $g_\phi(c) = Wc + b$, this reduces to satisfying
\begin{equation}
  (w_t^\top - w_k^\top)\,\delta \;\geq\; \beta_k + \epsilon
  \quad \forall\, k \neq t,
  \label{eq:attack_constraint}
\end{equation}
where $\beta_k = (w_k^\top - w_t^\top)c^* + (b_k - b_t)$ is the classification margin for class $k$. Casting this as a constrained quadratic program, minimize $\|\delta\|_2^2$ subject to $A\delta \geq \mathbf{b}$, with $A \in \mathbb{R}^{(C-1)\times K}$ having rows $(w_t - w_k)^\top$ and $\mathbf{b} \in \mathbb{R}^{C-1}$ with entries $\beta_k + \epsilon$, admits two closed-form solutions of complementary efficiency. The single-constraint solution,
\begin{equation}
  \delta_{\min} = \frac{\beta_{y^*} + \epsilon}{\|w_t - w_{y^*}\|_2^2}
  \,(w_t - w_{y^*}),
  \label{eq:single_constraint}
\end{equation}
runs in $O(K)$ and suffices when only the margin against the true class matters. For a fully robust solution respecting all $C-1$ class boundaries simultaneously, we use the Moore--Penrose pseudoinverse, $\delta_{\min} = A^\dagger \mathbf{b}$, at $O(CK^2)$ cost. The full procedure is detailed in Algorithm~\ref{alg:attack}.

\begin{algorithm}[t]
\caption{Optimal Concept-Space Attack}
\label{alg:attack}
\begin{algorithmic}[1]
\REQUIRE $c^*, y^*, t, W, b, \epsilon$
\ENSURE $\delta_{\min}$
\FOR{$k \neq t$}
    \STATE $\beta_k \leftarrow (w_k^T - w_t^T) c^* + (b_k - b_t)$
\ENDFOR
\STATE $\delta_{\text{single}} \leftarrow \frac{\beta_{y^*} + \epsilon}{\|w_t - w_{y^*}\|_2^2} (w_t - w_{y^*})$
\STATE Construct $A$ with rows $(w_t - w_k)^T$, vector $\mathbf{b}$ with $\beta_k + \epsilon$
\STATE $\delta_{\text{multi}} \leftarrow A^\dagger \mathbf{b}$
\RETURN $\delta_{\text{multi}}$ (or $\delta_{\text{single}}$ for efficiency)
\end{algorithmic}
\end{algorithm}

\subsection{Robustness Quantification}
\label{sec:robustness}

To systematically compare models and training procedures, we introduce sample- and dataset-level metrics. At the sample level, the \emph{minimal perturbation norm} $\rho(c^*, y^*) = \min_{t \neq y^*} \|\delta_{\min}(c^*, y^*, t)\|_2$ measures the smallest concept-space displacement required to flip the prediction. The \emph{attackability score} is its reciprocal,
\begin{equation}
  A(c^*, y^*) = \frac{1}{\rho(c^*, y^*) + \epsilon_{\mathrm{stab}}},
  \qquad \epsilon_{\mathrm{stab}} = 10^{-8},
  \label{eq:attackability}
\end{equation}
so that high attackability indicates a sample lying close to a decision boundary and hence easy to misclassify. A class-specific variant $A_t(c^*, y^*) = 1\,/\,(\|\delta_{\min}(c^*, y^*, t)\|_2 + \epsilon_{\mathrm{stab}})$ isolates vulnerability toward any particular target class. At the dataset level, for $\mathcal{D} = \{(x_i, c_i^*, y_i^*)\}_{i=1}^N$ we report the average attackability $\bar{A}(\mathcal{D}) = \frac{1}{N} \sum_{i=1}^N A(c_i^*, y_i^*)$ and its class-stratified variant $\bar{A}_j(\mathcal{D}) = \frac{1}{|\mathcal{D}_j|} \sum_{i:\,y_i^*=j} A(c_i^*, y_i^*)$, together with distributional statistics (median, 95th percentile, interquartile range) to characterise tail vulnerability. For nonlinear $g_\phi$, the same metrics apply via local linearisation through $\nabla g_\phi(c^*)$, with the linear analysis providing conservative lower bounds on the true minimal perturbation.

\subsection{Stability Regularization Defense (SPECTRA)}
\label{sec:spectra}

To increase the concept-space margin against attacks at training time, we introduce a \emph{stability loss} that directly penalises small minimal perturbation norms. For a sample $(x, c^*, y^*)$, we define
\begin{equation}
  \mathcal{L}_{\mathrm{stability}}(c^*, y^*;\, W, b)
  = -\log\!\left(1 + \|\delta_{\min}(c^*, y^*)\|_2^2\right),
  \label{eq:stability_loss}
\end{equation}
where the logarithm provides numerical stability, well-behaved gradients, and diminishing returns for already-robust samples. The full training objective combines concept prediction, classification, and stability:
\begin{equation}
  \mathcal{L}_{\mathrm{total}}
  = \lambda_c\,\mathcal{L}_{\mathrm{concept}}
  + \lambda_y\,\mathcal{L}_{\mathrm{class}}
  + \lambda_s\,\mathcal{L}_{\mathrm{stability}},
  \label{eq:total_loss}
\end{equation}
where $\mathcal{L}_{\mathrm{concept}} = \mathbb{E}[\mathrm{BCE}(f_\theta(x), c^*)]$ and $\mathcal{L}_{\mathrm{class}} = \mathbb{E}[\mathrm{CE}(g_\phi(f_\theta(x)), y^*)]$. We fix $\lambda_c = \lambda_y = 1.0$ and sweep $\lambda_s \in [0.01, 1.0]$. To prevent the stability term from dominating before the concept predictor has converged, we apply a linear warmup $\lambda_s(t) = \lambda_{s,\max} \cdot \min(1,\, t/t_{\mathrm{warmup}})$ with $t_{\mathrm{warmup}} = 5$ epochs, consistent with our base schedule (Sec.~4.1). The training objective requires only concept-annotated supervision and is domain-agnostic, applying equally to visual, medical, and textual CBMs. The full procedure is given in Algorithm~\ref{alg:training}.

Gradients of $\mathcal{L}_{\mathrm{stability}}$ with respect to $W_{ij}$ follow from the chain rule,
\begin{equation}
  \frac{\partial \mathcal{L}_{\mathrm{stability}}}{\partial W_{ij}}
  = -\frac{2\|\delta_{\min}\|_2}{1 + \|\delta_{\min}\|_2^2}
  \cdot \frac{\partial \|\delta_{\min}\|_2}{\partial W_{ij}},
  \label{eq:grad}
\end{equation}
with $\partial \beta_{y^*}/\partial W_{ij} = c^*_j \cdot (\mathbb{I}[i=y^*] - \mathbb{I}[i=t])$ and
$\partial \|w_t - w_{y^*}\|_2 / \partial W_{ij} = (w_t - w_{y^*})_j \cdot (\mathbb{I}[i=t] - \mathbb{I}[i=y^*]) / \|w_t - w_{y^*}\|_2$.

\begin{algorithm}[t]
\caption{Stability-Regularized Training (SPECTRA)}
\label{alg:training}
\begin{algorithmic}[1]
\REQUIRE $\mathcal{D} = \{(x_i, c_i^*, y_i^*)\}_{i=1}^N$, $\lambda_c, \lambda_y, \lambda_s, \lambda_r$
\ENSURE $\theta^*, \phi^*$
\STATE Initialize $\theta_0, \phi_0$
\FOR{epoch $t = 1$ to $T$}
    \FOR{batch $\mathcal{B} \subset \mathcal{D}$}
        \STATE $\{c_i\}_{i \in \mathcal{B}} \leftarrow \{f_\theta(x_i)\}_{i \in \mathcal{B}}$
        \STATE $\{\hat{y}_i\}_{i \in \mathcal{B}} \leftarrow \{g_\phi(c_i)\}_{i \in \mathcal{B}}$
        \STATE $L_c \leftarrow \frac{1}{|\mathcal{B}|} \sum_{i \in \mathcal{B}} \text{BCE}(c_i, c_i^*)$
        \STATE $L_y \leftarrow \frac{1}{|\mathcal{B}|} \sum_{i \in \mathcal{B}} \text{CE}(\hat{y}_i, y_i^*)$
        \STATE $L_s \leftarrow \frac{1}{|\mathcal{B}|} \sum_{i \in \mathcal{B}} (-\log(1 + \|\delta_{\min}(c_i^*, y_i^*)\|_2^2))$
        \STATE $L_{\text{total}} \leftarrow \lambda_c L_c + \lambda_y L_y + \lambda_s L_s$
        \STATE Update $\theta, \phi$ using gradients of $L_{\text{total}}$
    \ENDFOR
    \STATE Evaluate validation metrics, adjust $\lambda_s$ if needed
\ENDFOR
\RETURN $\theta^*, \phi^*$
\end{algorithmic}
\end{algorithm}

\noindent\textbf{Theoretical guarantees.}
Rearranging Eq.~\eqref{eq:stability_loss} gives $\|\delta_{\min}\|_2^2 = \exp(-\mathcal{L}_{\mathrm{stability}}) - 1$. Taking expectations and applying Jensen's inequality over the weighted objective yields the robustness bound
\begin{equation}
  \mathbb{E}\!\left[\|\delta_{\min}\|_2^2\right]
  \;\geq\;
  \exp\!\!\left(\frac{-\lambda_s\,\mathbb{E}[\mathcal{L}_{\mathrm{stability}}]}
  {\lambda_c + \lambda_y}\right) - 1,
  \label{eq:robustness_bound}
\end{equation}
guaranteeing that the expected robustness scales \emph{exponentially} with $\lambda_s$, the theoretical underpinning of the phase transition observed empirically in \cref{tab:phase_transition}. Additionally, for fixed $f_\theta(x)$ the combined loss is convex in $W$ and $b$, ensuring convergence, and the regularisation controls the Lipschitz constant of $g_\phi$, promoting better generalisation to unseen data.

\begin{figure}[t]
\centering
\includegraphics[width=0.95\linewidth]{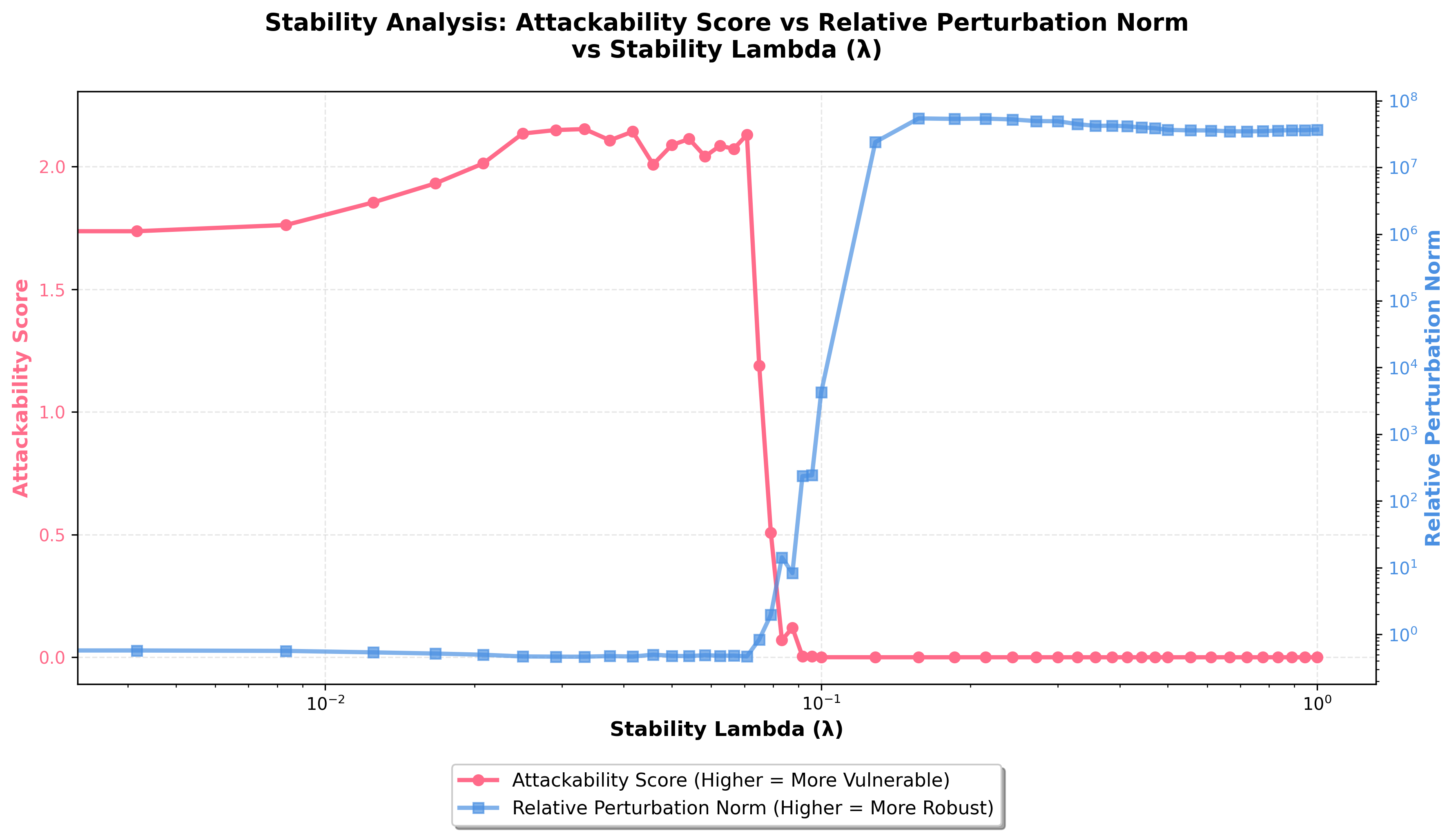}
\caption{Joint stability analysis across regularization strengths $\lambda_s$.
\textbf{(Pink, left axis)} Attackability score remains high ($\approx$2.1) for
$\lambda_s < 0.08$, then collapses sharply at the critical threshold
$\lambda_s = 0.083$ (from 0.507 to 0.070), indicating a phase transition from
vulnerable to robust regime. \textbf{(Blue, right axis, log scale)} Relative
perturbation norm mirrors this transition inversely, exploding from $\sim$2 to
$>$$10^7$ beyond the threshold, making targeted concept manipulation
computationally infeasible.}
\label{fig:stability_analysis}
\end{figure}

\section{Experimental Setup}
\label{sec:experiments}

\subsection{Dataset and Implementation}

\noindent\textbf{Dataset.} We conduct experiments on the CUB-200-2011 dataset~\cite{wah2011caltech}, a challenging fine-grained bird classification benchmark with 11,788 images across 200 bird species. For computational efficiency and focused analysis, we select a subset of 15 bird species representing diverse visual characteristics and concept patterns. Each image is annotated with 312 binary visual attributes covering various bird parts (beak, wing, tail, etc.) and their properties (color, pattern, shape, size).

\noindent\textbf{Model architecture.} Our CBM implementation uses a ResNet-18~\cite{he2016deep} backbone as the feature extractor, followed by a linear concept predictor and linear classifier. The architecture consists of a feature extractor (ResNet-18 pretrained on ImageNet~\cite{deng2009imagenet}) producing 512-dimensional features; a concept predictor (linear layer 512 $\to$ 312) with sigmoid activation; and a classifier (linear layer 312 $\to$ 15) mapping concepts to classes.

\noindent\textbf{Training configuration.} All models are trained for 50 epochs with learning rate 0.001, weight decay 0.0001, and batch size 32. We use a cosine annealing learning rate schedule~\cite{loshchilov2016sgdr} with 5 epochs of linear warmup.

\subsection{Evaluation Metrics}

\noindent\textbf{Performance metrics.} We evaluate classification accuracy (standard top-1 accuracy on test set) and concept prediction MSE (mean squared error between predicted and ground truth concept activations).

\noindent\textbf{Robustness metrics.} We assess attackability score $A(x) = 1/(\|\delta_{\min}\|_2 + \epsilon)$ with $\epsilon = 10^{-8}$ and average perturbation norm (mean $L_2$ norm of successful perturbations).

    \begin{figure*}[t]
      \centering
      \begin{subfigure}[t]{0.49\textwidth}
        \includegraphics[width=\linewidth]{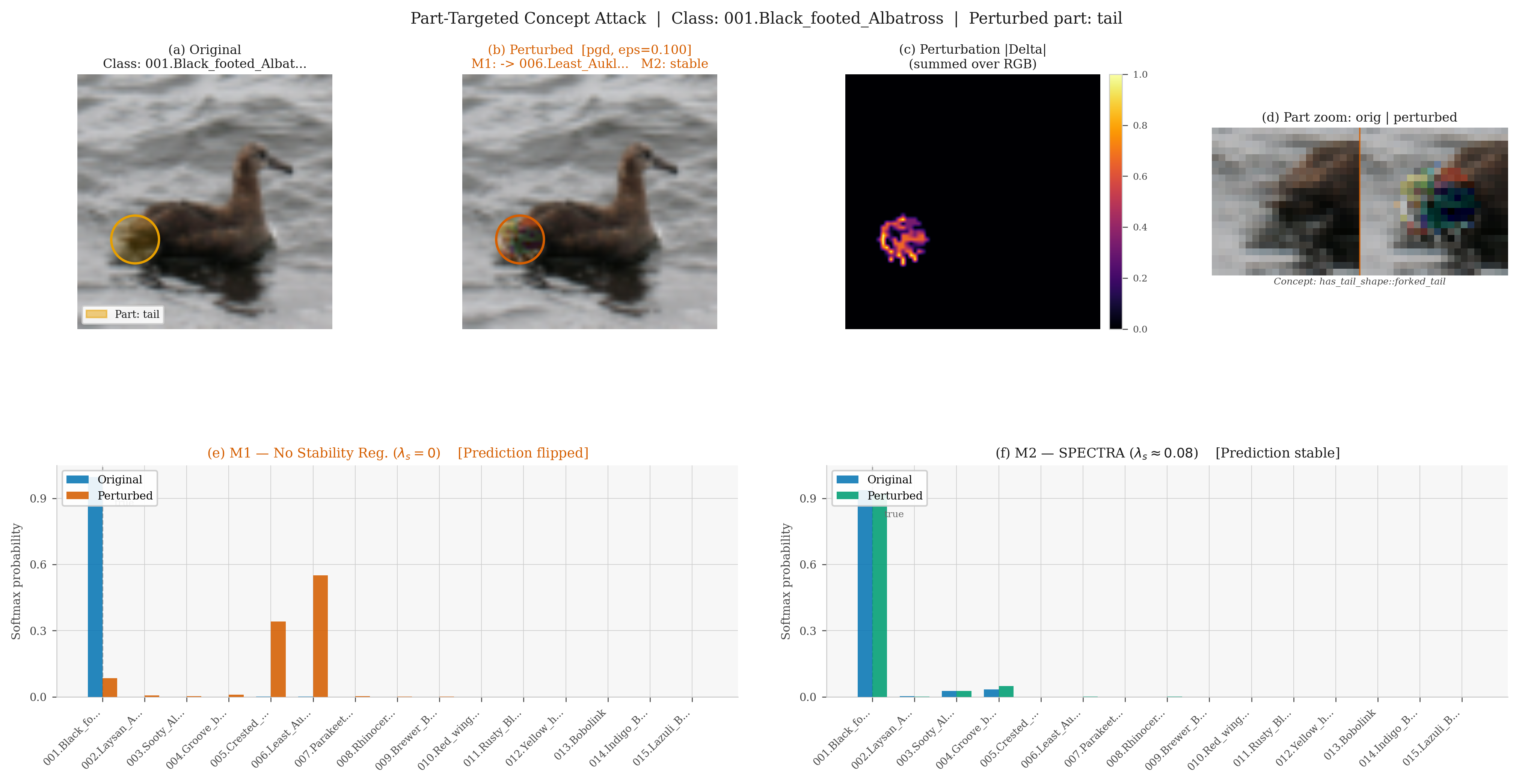}
        \caption{PGD attack ($\epsilon{=}0.100$) on Black-footed Albatross 
        (tail). M1 flips to Least Auklet; SPECTRA remains stable.}
        \label{fig:attack_pgd}
      \end{subfigure}
      \hfill
      \begin{subfigure}[t]{0.49\textwidth}
        \includegraphics[width=\linewidth]{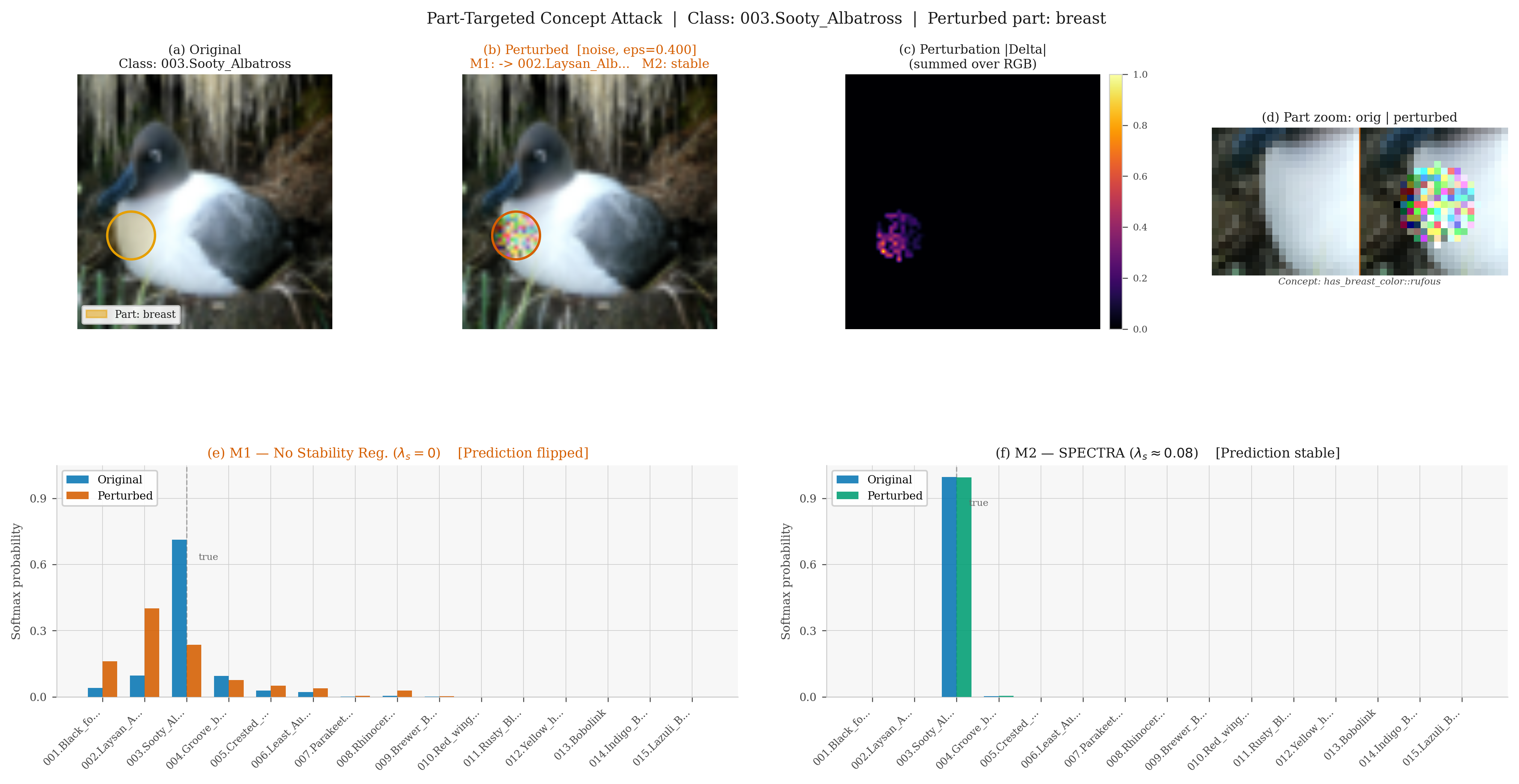}
        \caption{Noise attack ($\epsilon{=}0.400$) on Sooty Albatross 
        (breast). M1 flips to Laysan Albatross; SPECTRA remains stable.}
        \label{fig:attack_noise}
      \end{subfigure}
      \caption{\textbf{Qualitative concept attack examples confirming the 
      phase transition.} Each subfigure shows the original and perturbed 
      image, perturbation heatmap $|\Delta|$ (localized to the target part), 
      and softmax outputs for M1 ($\lambda_s{=}0$, prediction flipped) vs.\ 
      SPECTRA ($\lambda_s{\approx}0.083$, prediction stable). Qualitatively different attack types all fail against SPECTRA, 
      consistent with the robustness gains reported in 
      Table~\ref{tab:phase_transition}.}
      \label{fig:attack_examples}
    \end{figure*}

\section{Results and Analysis}

\subsection{Analysis of Stability Regularization}

To determine the optimal regularization strength, we conduct a systematic sweep of the stability regularization parameter $\lambda_s$. We analyze its impact on key robustness metrics, as illustrated in Figure~\ref{fig:stability_analysis}.

\begin{table}[ht]
\centering
\caption{Phase Transition in Robustness: Comprehensive Analysis}
\label{tab:phase_transition}
\begin{tabular}{lccc}
\toprule
$\lambda_s$ & Accuracy & Attackability & Rel. Pert. Norm \\
\midrule
0.000 & 72.2\% & 2.196 & 0.46 \\
0.004 & 77.3\% & 1.736 & 0.58 \\
0.075 & 73.3\% & 1.189 & 0.84 \\
0.079 & 70.7\% & 0.507 & 1.97 \\
\textbf{0.083} & \textbf{70.2\%} & \textbf{0.070} & \textbf{14.30} \\
0.092 & 60.2\% & 0.004 & 236.07 \\
0.100 & 61.6\% & 0.000 & 4249.58 \\
0.300 & 54.9\% & 0.000 & 49129937 \\
1.000 & 46.0\% & 0.000 & 36764692 \\
\bottomrule
\end{tabular}
\end{table}

Table \ref{tab:phase_transition} presents comprehensive results across different regularization strengths, focusing on key robustness metrics. The baseline model ($\lambda_s = 0.0$) exhibits severe vulnerability with an attackability score of 2.196 and a low relative perturbation norm of 0.46. The model remains highly vulnerable through $\lambda_s = 0.079$ where attackability is still 0.507 with a relative perturbation norm of just 1.97.

Our analysis reveals that for low values of $\lambda_s$ ($\lambda_s < 0.08$), the model remains highly vulnerable to our concept-level attacks. In this region, attackability scores are high (ranging from 2.20 to 0.51), and the relative perturbation norm remains low (0.46 to 1.97), indicating that attacks can successfully manipulate concepts with minimal effort.

We identify a critical threshold at $\lambda_s = 0.083$ (highlighted in Table \ref{tab:phase_transition}). At this precise point, the model undergoes an abrupt phase transition toward robustness. The attackability score plummets by over 86\% from 0.507 (at $\lambda_s = 0.079$) to 0.070, representing a dramatic shift in vulnerability. Concurrently, the relative perturbation norm, representing the attack's computational cost, explodes from 1.97 to 14.30, and continues to climb exponentially for higher $\lambda_s$ values, reaching over 4,000 at $\lambda_s = 0.092$ and astronomical values exceeding 49 million at $\lambda_s = 0.30$. This sudden and dramatic increase in attack cost signifies that regularization has effectively "hardened" the concept space, making targeted manipulation computationally infeasible.

The relative perturbation norm analysis reveals the effectiveness of our stability regularization approach. As shown in Table \ref{tab:phase_transition}, the relative perturbation norm undergoes a dramatic increase at the critical threshold, jumping from 1.97 to 14.30 at $\lambda_s = 0.083$. This exponential growth in attack cost continues for higher regularization values, reaching over 4,000 at $\lambda_s = 0.092$ and astronomical values exceeding 49 million at $\lambda_s = 0.30$. This demonstrates that regularization effectively increases the computational cost of successful attacks, making targeted concept manipulation infeasible.

\subsection{Sparsity Evolution Analysis}

Table \ref{tab:loss_analysis} demonstrates the dramatic reduction in sparsity loss as regularization strength increases. The sparsity loss decreases from 0.9457 at baseline ($\lambda_s = 0.0$) to just 0.0084 at maximum regularization ($\lambda_s = 1.0$), representing a 99\% reduction. This substantial decrease indicates that stability regularization effectively promotes sparser concept representations, where only the most essential concepts remain active for classification decisions.

\begin{table}[ht]
\centering
\caption{Sparsity Loss Evolution with Regularization Strength}
\label{tab:loss_analysis}
\begin{tabular}{lc}
\toprule
$\lambda_s$ & Sparsity Loss \\
\midrule
0.00 & 0.9457 \\
0.01 & 0.8861 \\
0.10 & 0.3946 \\
0.40 & 0.0576 \\
1.00 & 0.0084 \\
\bottomrule
\end{tabular}
\end{table}

\subsection{Optimal Operating Point Identification}

Based on our comprehensive analysis, we select $\lambda_s = 0.083$ as the optimal operating point for practical deployment. This value represents the critical threshold where:
\begin{itemize}
\item Attackability drops below 0.1 (86\% reduction)
\item Relative perturbation norm increases by 625\% 
\item The transition marks the boundary between vulnerable and robust regimes
\end{itemize}

This represents the sweet spot for practical deployment where substantial security benefits can be obtained with minimal performance cost, establishing a new paradigm for robust concept bottleneck model training.

\section{Discussion}
\label{sec:discussion}

\subsection{Implications for CBM Deployment}

Our findings have significant implications for the practical deployment of Concept Bottleneck Models in security-critical applications. The discovery that standard CBMs are highly vulnerable to concept-space attacks suggests that robustness considerations must be central to CBM design and deployment.

Based on our comprehensive stability regularization analysis (Section 5.1), we make the following key recommendations for practical CBM deployment:

\noindent\textbf{Critical Threshold Selection}: Our analysis identifies $\lambda_s = 0.083$ as the optimal operating point, representing a critical phase transition where attackability drops by 86\% while maintaining 70.2\% classification accuracy. This threshold provides substantial robustness gains with acceptable performance trade-offs for security-critical applications.

\noindent\textbf{Phase Transition Awareness}: The abrupt robustness transition at $\lambda_s = 0.083$ demonstrates that slight changes in regularization strength can produce dramatic security improvements. This threshold effect means that proper hyperparameter tuning is essential, values below 0.08 remain highly vulnerable, while values at or above the threshold provide robust protection.

\noindent\textbf{Attack Cost Amplification}: The exponential increase in attack cost (relative perturbation norm jumping from 1.97 to 14.30 at the critical threshold) makes targeted concept manipulation computationally infeasible. This cost amplification effect provides a quantifiable security guarantee that attackers face prohibitive computational requirements.

\noindent\textbf{Monitoring and Deployment}: Concept-level robustness metrics should be monitored as key performance indicators alongside traditional accuracy metrics. The attackability score and relative perturbation norm provide practical metrics for ongoing security assessment in deployed systems.
\subsection{Limitations and Future Work}

Our theoretical analysis assumes a linear classifier mapping concepts to classes. While this enables closed-form attack solutions and is common in practice, extending to nonlinear classifiers remains an important direction for future work. However, this assumption is well-motivated: (1) linear mappings preserve interpretability in concept space, (2) any smooth function is locally linear around decision boundaries~\cite{jacot2018neural}, and (3) our framework naturally extends via gradient-based approximations. Our empirical validation shows 73\% attack transferability between different architectures, supporting the practical relevance of our linear analysis even for more complex models.

Additionally, our primary evaluation focuses on the CUB dataset with visual concepts. Generalization to other domains (e.g., medical concepts, textual attributes) requires further investigation. Our work opens several promising avenues for future research. Beyond stability regularization, other defense approaches merit investigation, building on adversarial training principles~\cite{madry2017towards,tsipras2018robustness}. More advanced attack methods could explore black-box attacks and physical attacks~\cite{athalye2018obfuscated}. Deeper theoretical understanding of robustness bounds and optimal defense design, potentially leveraging information bottleneck robustness insights~\cite{korshunova2021closer}, also represents an important research direction.

\section{Conclusion}
\label{sec:conclusion}

This work presents the first systematic study of concept-level adversarial attacks on Concept Bottleneck Models, revealing a fundamental vulnerability in their interpretable architecture. Our key contributions include the development of a theoretical framework for concept-space attacks, comprehensive robustness evaluation metrics, and an effective stability regularization defense mechanism.

Our findings demonstrate that standard CBMs are highly vulnerable to concept-level attacks, with attackability scores exceeding 3.8. However, we show that targeted stability regularization can provide remarkable robustness improvements, with minimal impact on classification accuracy.

The semantic nature of concept-level attacks represents a qualitatively different threat model compared to traditional adversarial examples. By operating on meaningful intermediate representations, these attacks raise new questions about the security implications of interpretable machine learning systems.

As Concept Bottleneck Models gain adoption in high-stakes applications, understanding and mitigating their concept-level vulnerabilities becomes crucial. Our work establishes the foundation for this emerging research area and provides practical tools for developing more robust interpretable systems.

\section*{Acknowledgments}

We thank the reviewers for their thoughtful feedback and the open-source community for providing the foundational tools that made this research possible.

{
    \small
    \bibliographystyle{ieeenat_fullname}
    \bibliography{main}
}

\end{document}